\documentclass[conference,10pt]{IEEEtran}

\usepackage{setspace}
\usepackage{multicol}
\usepackage[final]{graphicx}
\usepackage{graphics}
\usepackage{amsmath,amsthm,amssymb,algorithmic, algorithm}
\usepackage{multicol}
\usepackage[margin=1in]{geometry}
\usepackage{subfig}
\usepackage{color}
\usepackage{setspace}
\usepackage{hyperref}
\usepackage{epstopdf}
\usepackage{soul}
\hypersetup{colorlinks=true}
\usepackage{mathtools}
\usepackage{microtype}
\usepackage{ifxetex,ifluatex}
\ifxetex
\usepackage{xltxtra}
\else
\ifluatex
\usepackage{luatextra}
\else
\usepackage{amssymb}
\usepackage[utf8]{inputenc}
\csname else\expandafter\endcsname
\romannumeral -`0
\fi
\fi
\iftrue\relax%
\defaultfontfeatures{Ligatures=TeX}
\usepackage[math-style=TeX, vargreek-shape=TeX]{unicode-math}
\fi

\hypersetup{colorlinks=true}

\usepackage{xparse}
\usepackage{tikz}
\usetikzlibrary{shapes.geometric, positioning, shadows}

\DeclareMathOperator{\rk}{rk} 
\DeclareMathOperator{\frco}{frc} 
\DeclareMathOperator{\rco}{rc} 

\theoremstyle{plain}

\NewDocumentCommand{\probability}{d()om}{%
  \operatorname{\mathbb{P}}%
  \IfValueT{#1}{\sb{#1}}%
  \left[#3%
    \IfValueT{#2}{\,\middle|\,#2}\right]}
\NewDocumentCommand{\expectation}{d()om}%
  {\operatorname{\mathbb{E}}%
      \IfValueT{#1}{\sb{#1}}%
      \left[#3%
    \IfValueT{#2}{\,\middle|\,#2}\right]}
\NewDocumentCommand{\entropy}{om}{\mathbb{H}\left[#2
    \IfValueT{#1}{\,\middle|\,#1}\right]}
\NewDocumentCommand{\bentropy}{lm}
  {\widetilde{\mathbb{H}}#1\left[#2\right]}

\NewDocumentCommand{\mutualInfo}{omm}{\mathbb{I}\left[#2;#3
    \IfValueT{#1}{\,\middle|\,#1}\right]}
\NewDocumentCommand{\privInfo}{omm}{\mathbb{W}\left[#2;#3
    \IfValueT{#1}{\,\middle|\,#1}\right]}
\DeclareExpandableDocumentCommand{\comm}{om}
  {\mathbb{C}\left[#2\IfValueT{#1}{\,\middle|\,#1}\right]}
\NewDocumentCommand{\priv}{om}
  {\mathbb{W}\left[#2\IfValueT{#1}{\,\middle|\,#1}\right]}
\NewDocumentCommand{\infrk}{om}
  {\rk_{\mathbb{I}}\left(#2\IfValueT{#1}{\,\middle|\,#1}\right)}

\NewDocumentCommand{\infset}{om}
  {Q\left(#2\IfValueT{#1}{\,\middle|\,#1}\right)}

\NewDocumentCommand{\frc}{om}
  {\frco\left(#2\IfValueT{#1}{\,\middle|\,#1}\right)}

\NewDocumentCommand{\rc}{om}
  {\rco\left(#2\IfValueT{#1}{\,\middle|\,#1}\right)}

\DeclareExpandableDocumentCommand{\dcomm}{om}
  {\mathbb{DC}\left[#2\IfValueT{#1}{\,\middle|\,#1}\right]}

\DeclareExpandableDocumentCommand{\rcomm}{om}
  {\mathbb{RC}\left[#2\IfValueT{#1}{\,\middle|\,#1}\right]}

\newtheorem{theorem}{Theorem}[section]
\newtheorem{lemma}[theorem]{Lemma}

\theoremstyle{definition}

\theoremstyle{remark}

\newtheoremstyle{claim}{}{}{}{}{\itshape}{.}{.5em}{}
\theoremstyle{claim}

  {\end{proof}}

\title{Sequential Low-Rank Change Detection}

\author{ \IEEEauthorblockN{Yao
Xie}
\IEEEauthorblockA{H. Milton Stewart School of Industrial \\and Systems Engineering\\
Georgia Institute of Technology, GA\\
Email: yao.xie@isye.gatech.edu}\and \IEEEauthorblockN{Lee Seversky}
\IEEEauthorblockA{Air Force Research Lab, \\Information Directorate, 
Rome, NY\\
Email: lee.seversky@us.af.mil} }

\date{\today}

\begin{document}


\maketitle
\begin{abstract}
Detecting emergence of a low-rank signal from high-dimensional data is an important problem arising from many applications such as camera surveillance and swarm monitoring using sensors. We consider a procedure based on the largest eigenvalue of the sample covariance matrix over a sliding window to detect the change. To achieve dimensionality reduction, we present a sketching-based approach for rank change detection using the low-dimensional linear sketches of the original high-dimensional observations. The premise is that when the sketching matrix is a random Gaussian matrix, and the dimension of the sketching vector is sufficiently large, the rank of sample covariance matrix for these sketches equals the rank of the original sample covariance matrix with high probability. Hence, we may be able to detect the low-rank change using sample covariance matrices of the sketches without having to recover the original covariance matrix. We character the performance of the largest eigenvalue statistic in terms of the false-alarm-rate and the expected detection delay, and present an efficient online implementation via subspace tracking.
\end{abstract}

\section{Introduction}
\label{sec:introduction}

Detecting emergence of a low-rank structure is a problem arising from many high-dimensional streaming data applications, such as video surveillance, financial time series, and sensor networks. The subspace structure may represent an anomaly or novelty that we would like to detect as soon as possible once it appears. One such example is swarm behavior monitoring. Biological swarms consist of many simple individuals following basic rules to form complex collective behaviors \cite{Berger16}. Examples include flocks of birds, schools of fish, and colonies of bacteria. The collective behavior and movement patterns of swarms have inspired much recent research into designing robotic swarms consisting of many agents that use simple algorithms to collectively accomplish complicated tasks, e.g., a swarm of UAVs \cite{swarm_drone16}. 
Early detection of an emerging or transient behavior that leads to a specific form of behavior is very important for applications in swarm monitoring and control. One key observation, as shown in \cite{Berger16} for classification of behavior purposes, is that many forms of swarm behaviors are represented as low-dimensional linear subspaces. This leads to a low-rank change in the covariance structure of the observations.

\begin{figure}[h!]
\begin{center}
\includegraphics[width = 0.3\textwidth]{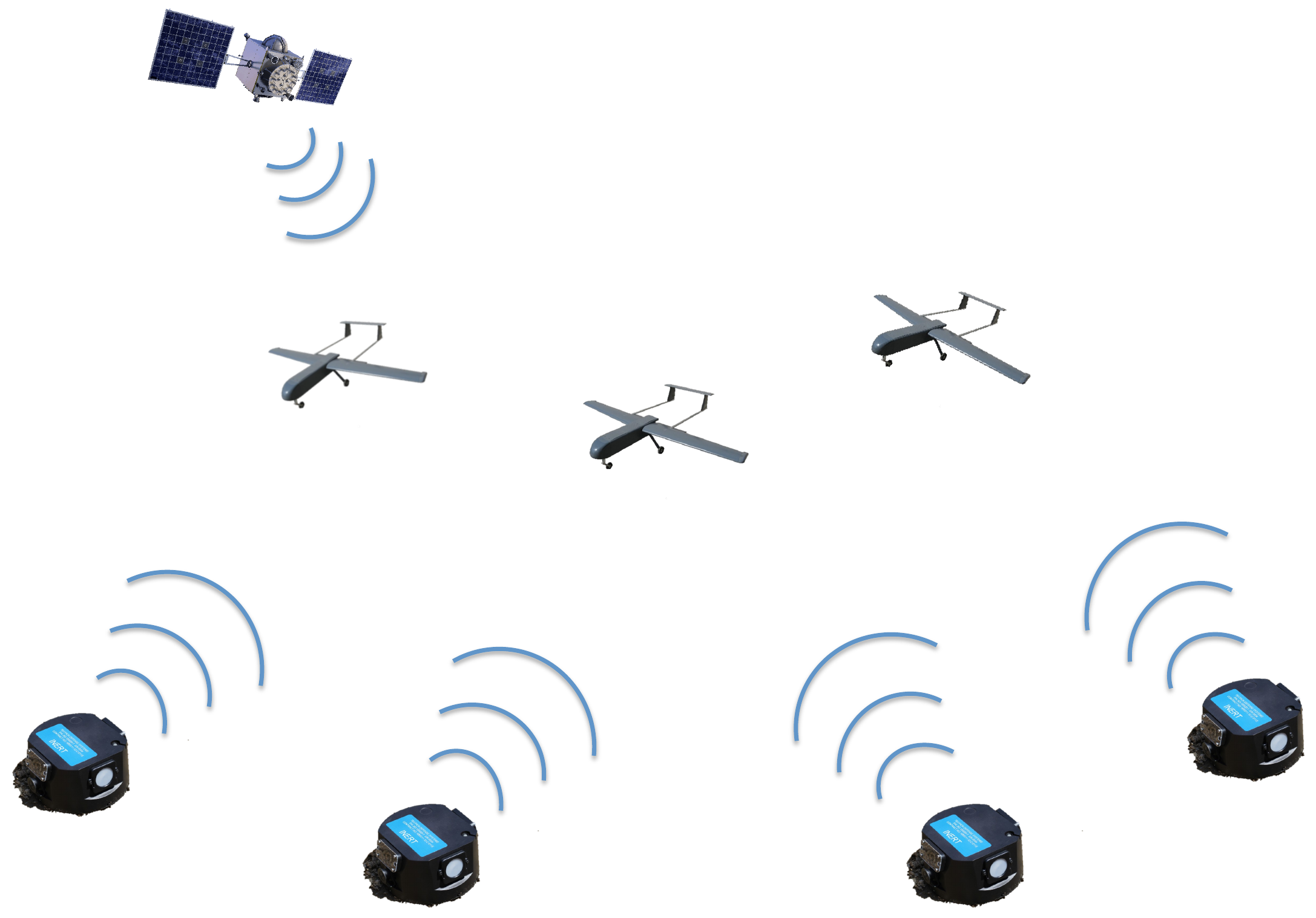} 
\caption{
Sensors  jointly monitor locations (and possibly speeds) of a swarm. }
\end{center}
\label{fig:sensing_scene}
\end{figure}

\begin{figure}[h!]
\begin{center}
\includegraphics[width = 0.4\textwidth]{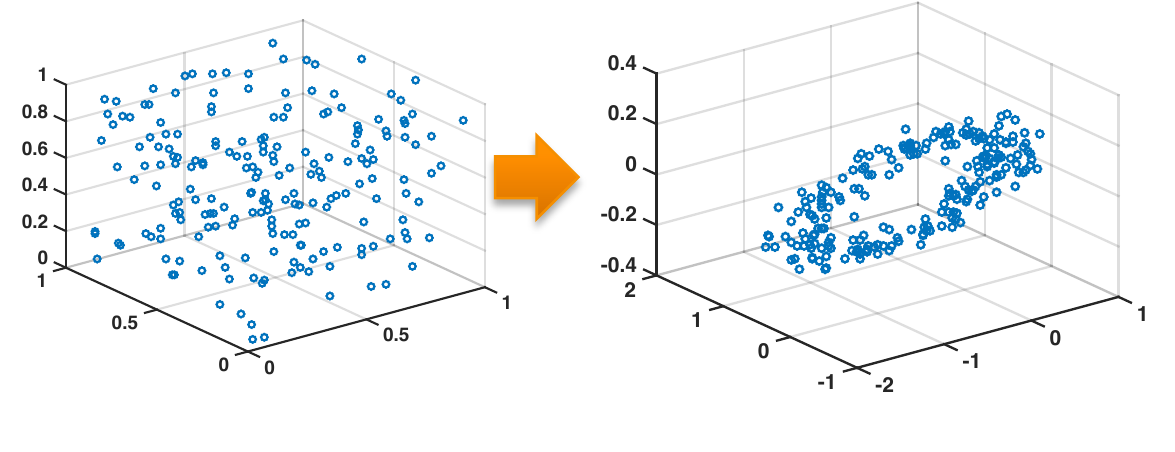}
\caption{Detection of emergence of a low-rank signal. The observations are the locations of the particles. Left, locations of the particles are random and the sample covariance matrix is due to noise. Right, the locations of the swarms lie near a circle in the three dimensional space, and the sample covariance exhibit of a strong rank-two component.  
}
\end{center}
\label{fig:demo}
\end{figure}

In this paper, we propose a sequential change-point detection procedure based on the maximum eigenvalue of the sample covariance, which is a natural approach to detect the emergence of a low-rank signal. We characterize the performance of the detection procedure using two standard performance metrics, the average-run-length (ARL) that is related to the false-alarm-rate, and the expected detection delay.  The distribution of the largest eigenvalue of the sample covariance matrix when data are Gaussian is presented in \cite{Johnstone01}, which corresponds to the Tracey-Widom law of order one. However, it is intractable to directly analyze ARL using the Tracey-Widom law. Instead, we use a simple $\varepsilon$-set argument to decompose the detection statistic into a set of  $\chi^2$-CUSUM procedures, whose performance is well understood in literature. 

Moreover, we present a new approach to detect low-rank changes based on sketching of the observation vectors. Sketching corresponds to linear projection of the original observations: $y_t = A x_t$, with $A \in \mathbb{R}^{m\times p}$. We may use sample covariance matrix of the sketches to detect emergence of a low-rank signal component, since it can be shown that for random projection, it can be shown that when $m$ is greater the rank of the signal covariance matrix, the sample covariance matrix of the linear sketches will have the same rank as the signal covariance matrix. We show that the minimum number of sketches $m$ is related to the property of the projection $A$ and the signal subspace. One key difference of the sketching for low-rank change detection from covariance sketching for low-rank covariance matrix recovery \cite{ChenChi15} is that, here we do not have to recover the covariance matrix which may require more number of sketches. Hence, sketching is a natural fit to detection problem. 

The low-rank change is related to the so-called spiked covariance matrix \cite{Johnstone01}, which assumes that a small number directions explain most of the variance. Such assumption is also made by sparse PCA, where the low-rank component is further assumed to be sparse. The goal of sparse PCA is to estimate such sparse subspaces (see, e.g., \cite{CaiMaWu16}). A fixed-sample hypothesis test to determine whether or not there exists a sparse and low-rank component, based on the largest eigenvalue statistic, is studied in \cite{Berthet13}, where it is shown to asymptotically minimax optimal. Another test statistic for such problems, the so-called Kac-Rice statistic, has been studied in \cite{TaylorRice}. The Kac-Rice statistic is the conditional survival function of the largest observed singular value conditioned on all other observed singular values, and it has a simple distribution form as uniformly distributed on [0, 1]. However, the statistic involves an infinite integral over the real line, which may not be easy to evaluate, and the test statistic needs to compute all eigenvalues of the sample covariance matrix instead of the largest eigenvalue.

\section{largest eigenvalue procedure}

Assuming a sequence of $p$-dimensional vectors $x_1, x_2, \ldots, x_t$, $t = 1, 2, \ldots$. There may be a change-point at time $\tau$ such that the distribution of the data stream changes. Our goal is to detect such a change as quickly as possible. Formally, such a problem can be stated as the following hypothesis test: 
\begin{align*}
H_0:  &\quad x_1, x_2, \ldots, x_t \stackrel{iid}{\sim} \mathcal{N}(0, \sigma_0^2 I_p) \\
H_1: & \quad
x_1, x_2, \ldots, x_\tau \stackrel{iid}{\sim} \mathcal{N}(0, \sigma_0^2 I_p),  \\
&~~~x_{\tau+1}, \ldots, x_t, \stackrel{iid}{\sim} \mathcal{N}(0, \sigma_0^2 I_p + \Sigma).
\end{align*}
Here $\sigma_0^2$ is the noise variance, which is assumed to be known or has been estimated from training data. Assume the signal covariance matrix $\Sigma$ is low-rank, meaning $\mbox{rank}(\Sigma) < p$. The signal covariance matrix is unknown.

One may construct a maximum likelihood ratio statistic. However, since the signal covariance matrix is unknown, we may have to form the generalized likelihood ratio statistic, which replaces the covariance matrix with the sample covariance. This may cause an issue since the statistic involves inversion of the sample covariance matrix, whose numerical property (such as condition number) is usually poor when $p$ is large.

Alternatively, we consider the largest eigenvalue of the sample covariance matrix which is a natural detection statistic for detecting a low-rank signal. Assume a scanning window approach. We estimate the sample covariance matrix using samples in a time window of $[t-w, t]$, where $w$ is the window size. Assume $w$ is chosen sufficiently large and it is greater than the anticipated longest detection delay. Using the ratio of the largest eigenvalue of the sample covariance matrix relative to the noise variance, we form the {\it maximum eigenvalue procedure} which is a stopping time given by:
\[
T = \inf\{t: \max_{t-w < k<t} (t-k)\left[\frac{1}{\sigma_0^2}\lambda_1(\widehat{\Sigma}_{t,k}) - d\right] \geq b \}, 
\]
where $d$ is the pre-set drift parameter, $b > 0$ is the threshold, and $\lambda_1(\Sigma)$ denotes the largest eigenvalue of a matrix $\Sigma$. 
Here the index $k$ represents the possible change-point location. Hence, samples between $[k+1, t]$ corresponds to post-change samples. The sample covariance matrix for post-change samples up to time $t$ is given by: \[\widehat{\Sigma}_{t, k} = \frac{1}{t-k} \sum_{i=k+1}^t x_t x_t^\intercal.\]
The maximization over $k$ corresponds to search for the unknown change-point location. An alarm is fired whenever the detection statistic exceeds the threshold $b$. 

\section{Performance bounds}\label{sec:bounds}

In this section, we characterize the performance of the procedure in terms of two standard performance metrics, the expected value of the stopping time when there is no change, called the average run length (ARL), and the expected detection (EDD), which is the expected time to stop in the extreme case when the change occurs immediately at $\kappa = 0$. 

Our approach is to decompose the maximum eigenvalue statistic to a set of $\chi^2$ CUSUM procedures. Our argument is
based on the $\varepsilon$-net, which provides a convenient way to discretize unit sphere in our case. The number of such compact set is called the covering number, $\mathcal{C}(X, \varepsilon)$, which is the minimal cardinality of an $\varepsilon$-net. The covering number of a unit sphere is given by 
\begin{lemma}[Lemma 5.2, \cite{Vershynin2011}]\label{lemma1}
The unit Euclidean sphere $S^{p-1}$ equipped with the Euclidean metric satisfies for every $\varepsilon > 0$ that 
\[
\mathcal{C}(S^{p-1}, \varepsilon) \leq \left(1+ \frac{2}{\varepsilon}\right)^p
\]
\end{lemma}

The $\varepsilon$-net help to derive an upper bound of the largest eigenvalue, as stated in the following lemma
\begin{lemma}[Lemma 5.4, \cite{Vershynin2011}]\label{lemma2}
Let $\Sigma$ be a symmetric $p\times p$ matrix, and let $\mathcal{C}_\varepsilon$ be an $\varepsilon$-net of $S^{p-1}$ for some $\varepsilon \in  [0, 1/2)$. Then 
\[
\lambda_1(A) \leq (1-2\varepsilon)^{-1} \sup_{q \in \mathcal{C}_\varepsilon} |q^\intercal A q|
\]
\end{lemma}

Our main theoretical results are the following, which are the lower bound on the ARL and the approximation to EDD of the largest eigenvalue procedure. 
\begin{theorem}[Lower bound on ARL]
When $b\rightarrow \infty$
\[
\mathbb{E}^\infty[T] \gtrsim \frac{ e^{b(1/2 - \varepsilon)(1-\theta)} }{(\frac{1-\theta}{2} + \frac{1}{2}\log \theta)(1+\frac{2}{\varepsilon})^p}.
\]
where $\theta\in (0, 1)$ is the root to the equation $\log \theta + d(1-\theta)(1-2\varepsilon) = 0$, $\varepsilon \in (0, \frac{1}{2})$. 
\end{theorem}
The lower bound above can be used to control false-alarm of the procedure. Given a target ARL, we may choose the corresponding threshold $b$. 

Let $\|\Sigma\|$ denote the spectral norm of a matrix $\Sigma$, which corresponds to the largest eigenvalue. We have the following approximation: 
\begin{theorem}[Approximation to EDD]
When $b\rightarrow \infty$
\begin{equation}
\mathbb{E}^1[T] = \frac{b+ e^{-b} -1}{[2(1 + \rho^2/\sigma_0^2)]^{-1} + \frac{1}{2}\log\left(1+\rho/\sigma_0\right)}(1+o(1))\label{EDD}
\end{equation}
where $\rho^2 = \|\Sigma\|$.
\end{theorem}
In (\ref{EDD}), $\rho^2/\sigma^2$ represents the signal-to-noise ratio. 
Note that the right-hand-side of (\ref{EDD}) is a decreasing function of $\rho^2/\sigma^2$, which is expected, since the detection delay should be smaller when the signal-to-noise ratio is larger.

The following informal derivation justifies the theorems. First, from Lemma \ref{lemma1}, we have that the detection statistic
\[
\lambda_1(\widehat{\Sigma}_{t,k})
\leq (1-2\varepsilon)^{-1} \max_{q\in \mathcal{C}_\varepsilon}
|q^\intercal \widehat{\Sigma}_{t, k} q|
\]
For each $q\in \mathcal{C}_\varepsilon(S)$, $\|q\| = 1$, we have
\[
\frac{(t-k)}{\sigma_0^2}|q^\intercal \widehat{\Sigma}_{t, k} q|
=  \sum_{i=k+1}^t \frac{(q^\intercal x_i)^2}{\sigma_0^2}
\]
Note that under $H_0$: $x_i \sim \mathcal{N}(0, \sigma_0^2 I_p)$, and hence $q^\intercal x_i/\sigma_0 \sim \mathcal{N}(0, 1)$, and
\[(q^\intercal x_i)^2/\sigma_0^2 \sim \chi^2(1).\]
Alternatively, under $H_1$: $x_i \sim \mathcal{N}(0, \sigma_0^2 I_p + \Sigma)$, and hence $q^\intercal x_i/\sigma_0 \sim \mathcal{N}(0, 1+q^\intercal \Sigma q/\sigma^2)$, and hence
\[(q^\intercal x_i)^2/\sigma_0^2 \sim (1+\frac{q^\intercal \Sigma q}{\sigma_0^2})\chi^2(1).\]
Hence, the distribution before the change is $\chi^2$ random variable, and after the change-point is a scaled $\chi^2$ random variable. Now
define a set of procedures, for each $q \in \mathcal{C}_{\varepsilon}(S)$:
\[
\widetilde{T}_q = \inf\{t:   \max_{k<t}\sum_{i=k+1}^t ((q^\intercal x_i)^2/\sigma_0^2 - w') \geq b'\}
\]
Note that each one of these procedures is a $\chi^2$ CUSUM procedure. Now if set
\[
w' = (1-2\varepsilon)w,\quad b'=(1-2\varepsilon)b.
\]
then due to the above relations, we may bound the average run length of the original procedure in terms of these $\chi^2$ procedures:
\[
\mathbb{E}^\infty[T] \geq \mathbb{E}^\infty\left[\min_q \widetilde{T}_q\right].
\]
Since each $\widetilde{T}_q$ is a $\chi^2$ CUSUM procedure, whose properties are well understood, we may obtain a lower bound to the ARL of the maximum eigenvalue procedure. 

\begin{figure}
\centering
\includegraphics[width = 0.25\textwidth]{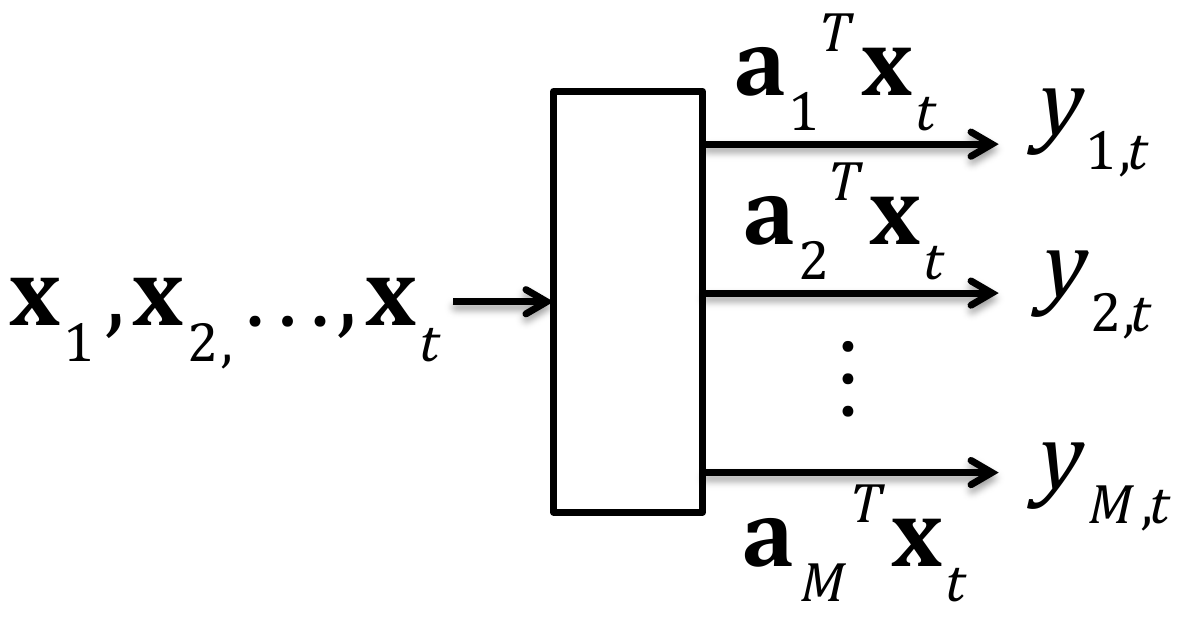}
\caption{Obtain $M$ sketches of each $n$-dimensional signal vector.}
\label{sketch}
\end{figure}

\section{Sketching for rank change detection}\label{indirect}

When $p$ is large, a common practice is to use a linear projection $A$ to reduce data dimensionality. The linear projection maps the original $p$-dimensional vector into a lower dimension $m$-dimensional vector. We refer such linear projection as sketching. One implementation of sketching is illustrated in Fig. \ref{sketch}, where each signal vector $x_t$ is projected by $M$ vectors, $a_1$, \ldots, $a_M$. The sketching corresponds to linear projection of the original vector: 
\[
y_{it} = a_i^\intercal x_t.
\]
Define a vector of observations $y_t = [y_{1, t}, \ldots, y_{M, t}]^\intercal \in \mathbb{R}^M$, and 
\[
A = [a_1, \ldots, a_M] \in \mathbb{R}^{p\times M}.
\]
we have $y_t = A^\intercal x_t$, $t = 1, 2, \ldots$.

One intriguing question is whether we may perform detection of the low-rank change using the linear projections. The answer is yes, as we present in the following. We first show that each linear corresponds to a {\it bi-linear} projection of the original covariance matrix. Define the sample covariance matrix of the sketches
\begin{equation}
\begin{split}
\widehat{\Sigma}^y_{k,t} & = \frac{1}{t-k} \sum_{i=k+1}^t y_t y_t^\intercal \\
& = \frac{1}{t-k}  \sum_{i=k+1}^t \begin{bmatrix}
y_{1i}^2 & y_{1i} y_{2i} & \cdots & y_{1i} y_{Mi} \\
y_{2i} y_{1i} & y_{2i}^2 & \cdots &  \\
\vdots & \ddots & \\
y_{Mi} y_{1i} & y_{Mi} y_{2i} & \cdots & y_{Mi}^2 
\end{bmatrix} \\
& =\frac{1}{t-k}  \sum_{i=k+1}^t 
\begin{bmatrix}
a_{1}x_i x_i^\intercal a_1 & a_1x_{i} x_{i} a_2 & \cdots & a_1x_{i} x_{i} a_M \\
a_2 x_{i} x_{i} a_1 & a_2 x_{i} x_{i} a_2 & \cdots &  \\
\vdots & \ddots & \\
a_M x_{i} x_{i} a_1 & a_M x_{i} x_{i} a_2 & \cdots & a_M x_{i} x_{i} a_M 
\end{bmatrix} \\
& = A^\intercal \widehat{\Sigma}_{t,k} A
\end{split}
\label{1}
\end{equation}

A key observation is that for certain choice of $A$, e.g., Gaussian random matrix, when $M >\mbox{rank}(\Sigma)$, the rank of $\widehat{\Sigma}_y$ is equal to the rank of $\widehat{\Sigma}_x$ with probability one. Hence, we may detect change using the largest eigenvalue of the sample covariance matrix of $y_t$.

A related line of work is {\it covariance sketching}, where the goal is to recover a low-rank covariance matrix using the {\it quadratic sketches}, which are square of each $(a_i^\intercal x_t)^2$. This corresponds to only using {\it diagonal entries} of the sample covariance matrix (\ref{1}) of the sketches.


\begin{figure}
\centering
\includegraphics[width = .4\textwidth]{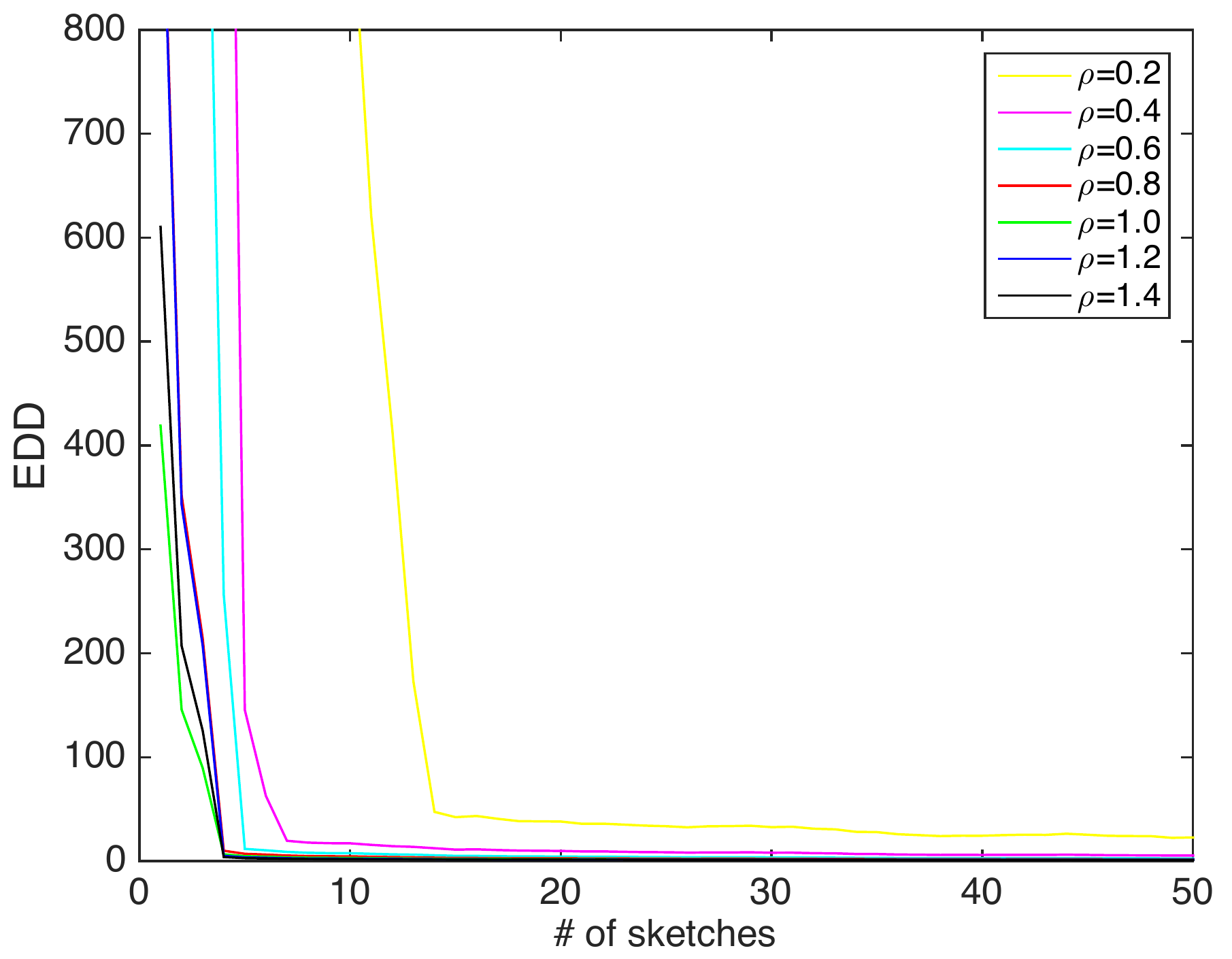}
\includegraphics[width = .4\textwidth]{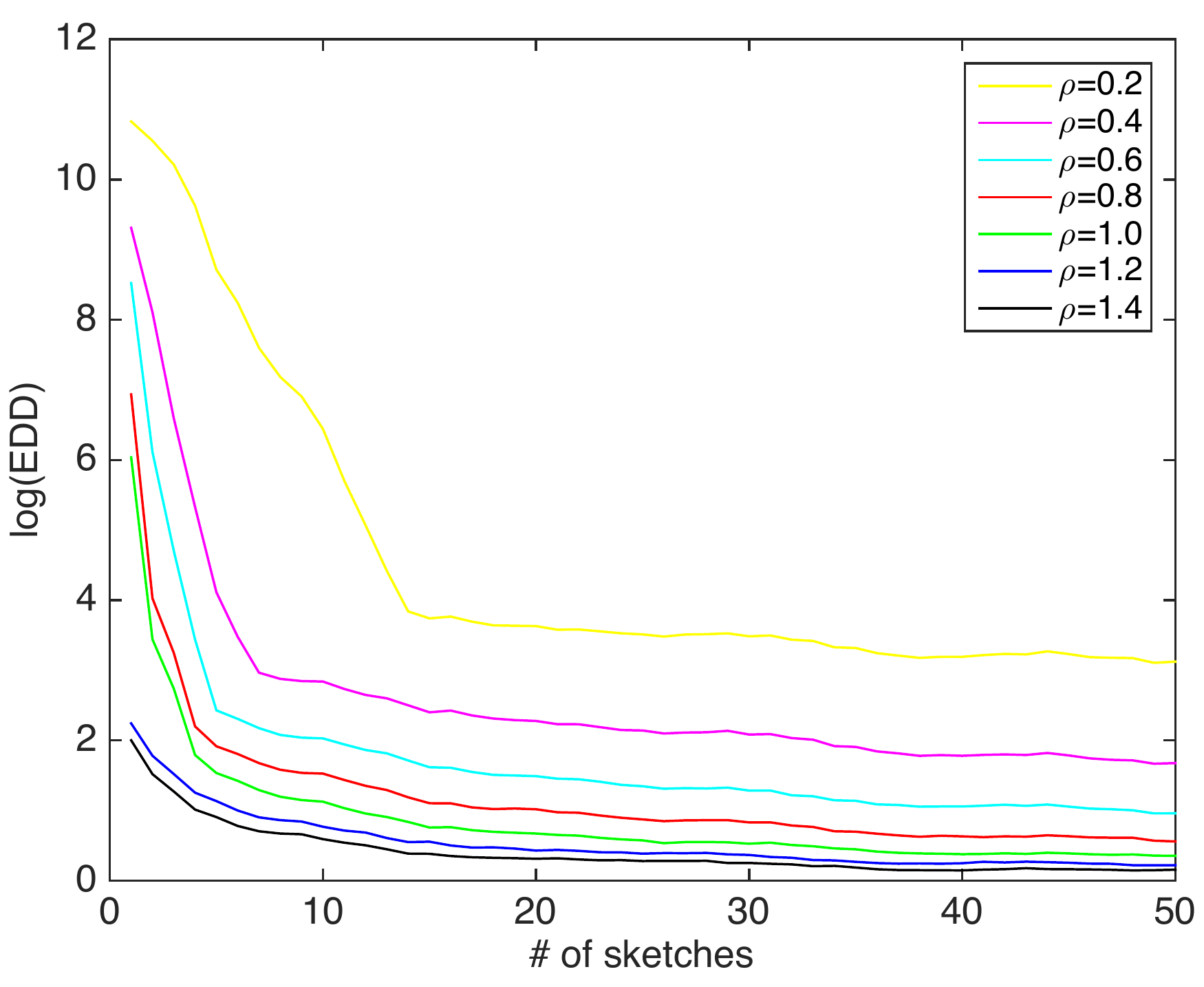}
\caption{EDD (upper) and logarithm of EDD (lower) versus $M$ (the number of sketches), for various SNR by varying $\rho$. Here $p = 100$, and $\sigma_0^2 = 1$, and the post-change covariance matrix is generated as 
 $\sigma_0^2I_m + \rho \times$ Gaussian random matrix with rank 3.}
\label{result}
\end{figure}

The sketches $y_t$ are still multi-variate Gaussian distributed, and the projection changes the variance. For the sketches, we may consider the following hypothesis test based on the original test form: 
\begin{align*}
H_0:  &\quad y_1, y_2, \ldots, y_t \stackrel{iid}{\sim} \mathcal{N}(0, \sigma_0^2 A^\intercal A) \\
H_1: & \quad
y_1, y_2, \ldots, y_\tau \stackrel{iid}{\sim} \mathcal{N}(0, \sigma_0^2 A^\intercal A),  \\
&~~y_{\tau+1}, \ldots, y_t, \stackrel{iid}{\sim} \mathcal{N}(0, \sigma_0^2 A^\intercal A + A^\intercal \Sigma A)
\end{align*}
Without loss of generality, to preserve noise property (or equivalently, to avoid amplifying noise), i.e., we choose the projection to be orthogonal, i.e.,
 $A ^\intercal A = I_m$. Moreover, due to the Gaussian form, the analysis for the ARL and EDD of the procedure in Section \ref{sec:bounds} still holds, with $\rho$ refined by $\|A^\intercal \Sigma A\|$. Hence, it can be seen from this analysis, that projection affects the signal-to-noise ratio, and hence, not surprisingly, although in principal we only need $M$ to be greater than the rank of $\Sigma$, in fact, we cannot choose $M$ to be arbitrarily small. 
 
Fig. \ref{result} illustrates this effect. In this example, we use a random subspace $A$ of dimension $M$ by $p$, and vary the number of sketches $M$, when the post-change covariance matrix of $x_t$ is generated as $\sigma_0^2 I_n$ plus a Gaussian random matrix with rank 3. Then we plot the EDD of the largest eigenvalue procedure when sketches are used. 
We calibrate the thresholds in each setting so that the ARL is fixed to be 5000.  
Note that when the number of sketches is sufficiently large (and greater than $\mbox{rank}(\Sigma)$, the EDD approaches to a small number. 
 
The following analysis may shed some light on how small we may choose $M$ to be. Let $s = \mbox{rank}(\Sigma)$, which is smaller than the ambient dimension $p$. Since the EDD (\ref{EDD}) is a decreasing function of $\mbox{SNR}$, which is a proportion to $\|A^\intercal \Sigma A\|$. Denote the eigendecomposition of $\Sigma$ to be $U\Lambda U^\intercal$. Clearly, SNR for the sketching case will depend on $A^\intercal U$, which depends on the principal angle between two subspaces. Recall that we have required $A$ to be a subspace $A^\intercal A = I_m$, and hence the SNR will depend on the principal angle between the random projection and the signal subspace $U$. 
The principal angle between two random subspaces is studied in \cite{AbsilEdelmanKoev06}. The eigenvalues of the sample covariance matrix are jointly Beta distribution. Based on this fact, the following lemma is obtained in \cite{CaoThompson15}, which helps us to understand the behavior of the procedure. It characterizes the $\ell_2$ norm of the fixed unit-norm vector projected by a random subspace. 
\begin{lemma}[Theorem 3 \cite{CaoThompson15}]
Given random subspace $A$, then for any fixed vector $u$ with $\|u\| = 1$, $\|A^\intercal u\| \sim \mbox{Beta}(m/2, (p-m)/2)$, and when $p\rightarrow \infty$ with $\delta = \lim_{p\rightarrow \infty} m/p$, for $0 < \varepsilon < \min(\delta, 1-\delta)$, \[
\mathbb{P}(\delta - \varepsilon < \|A^\intercal u\| < \delta + \varepsilon)\rightarrow 1
\]
\end{lemma}
Using this lemma in our setting, we may obtain a simple lower bound for the $\|A^\intercal U\|$:
\begin{equation}
\begin{split}
\|A^\intercal U\| &= \|[A^\intercal u_1 \cdots A^\intercal u_s]\|  = \max_{z: \|z\| = 1} \left\|\sum_{i} A^\intercal u_i z_i\right\|\\
&\geq \max_{i=1}^s \|A_i^\intercal u_i\| \gtrsim \frac{M}{p}\log s,
\end{split}
\label{lower_bound_subspace}
\end{equation}
with high probability, where the last inequality is due to the maximum of a set of Beta random variables. Hence, the number of sketches $M$ should scale as $p/\log s$. 

\begin{figure}
\begin{center}
\includegraphics[width =0.7\linewidth]{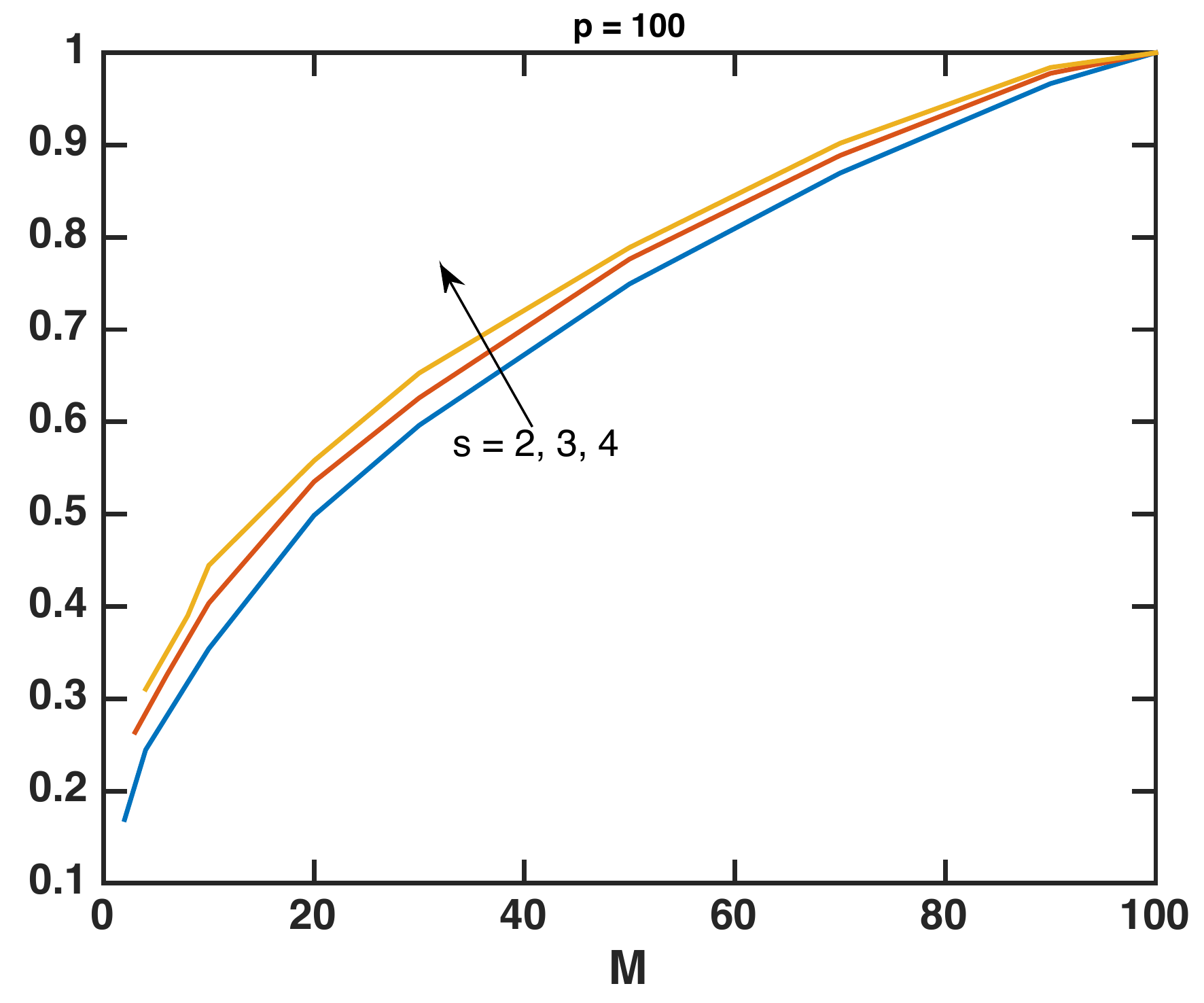}
\caption{The spectral norm $\|A^\intercal U\|$ versus $M$, when the number of sketches $M$ increases. Here $A$ is a subspace of dimension $p$-by-$m$, and $U$ is a orthonormal matrix of dimension $p$-by-$s$. Note that $\|A^\intercal U\|$ increases slowly as $s$ increases, and it also increases when $M$ increases, which is consistent with our theory in (\ref{lower_bound_subspace}).}
\end{center}
\end{figure}

\section{Online implementation via subspace tracking}\label{subspace}

\begin{figure}[h!]
\centering
\includegraphics[width =0.7\linewidth]{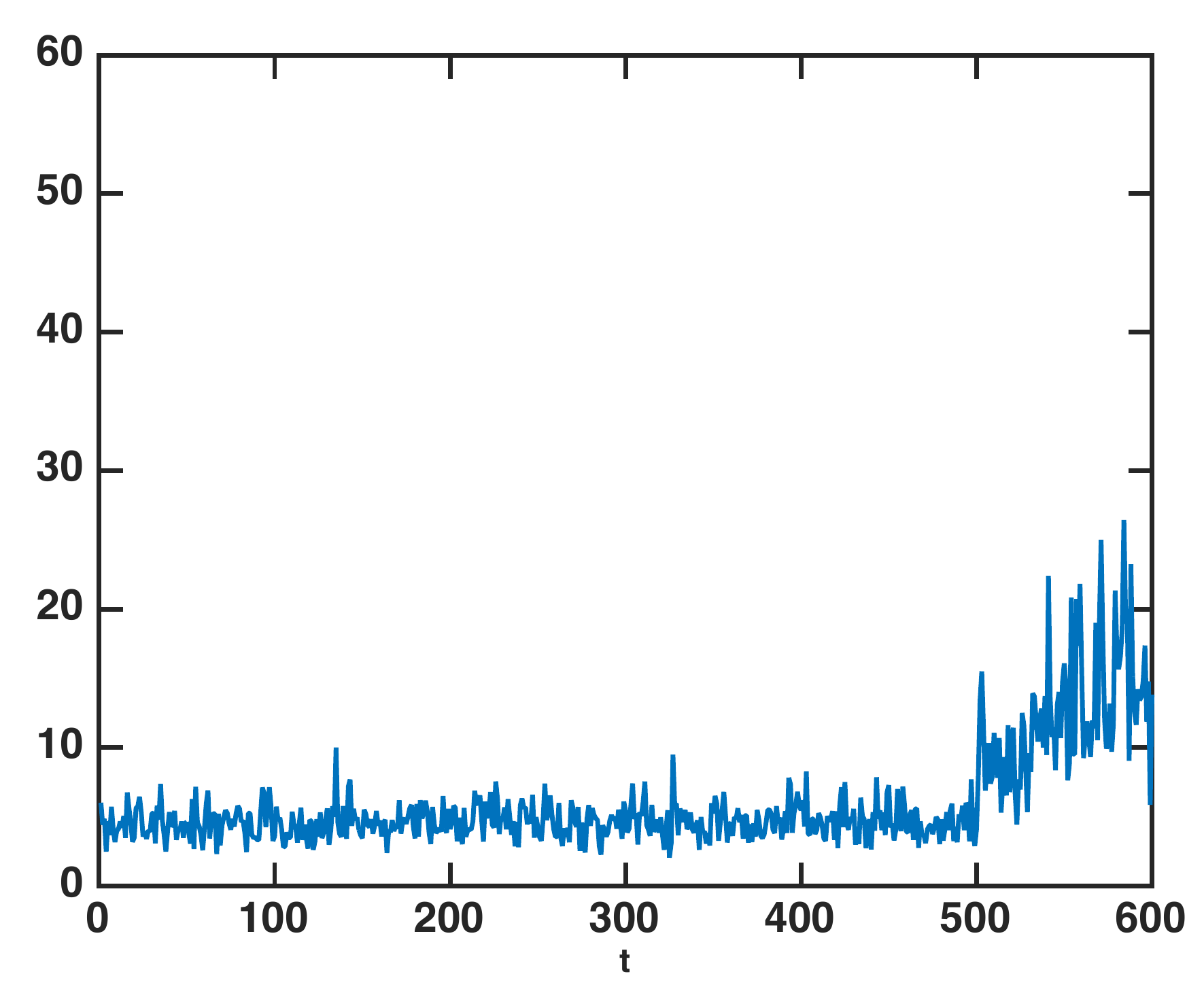} \\
(a): $\max|[\beta_t]_i|$\\
\includegraphics[width = 0.75\linewidth]{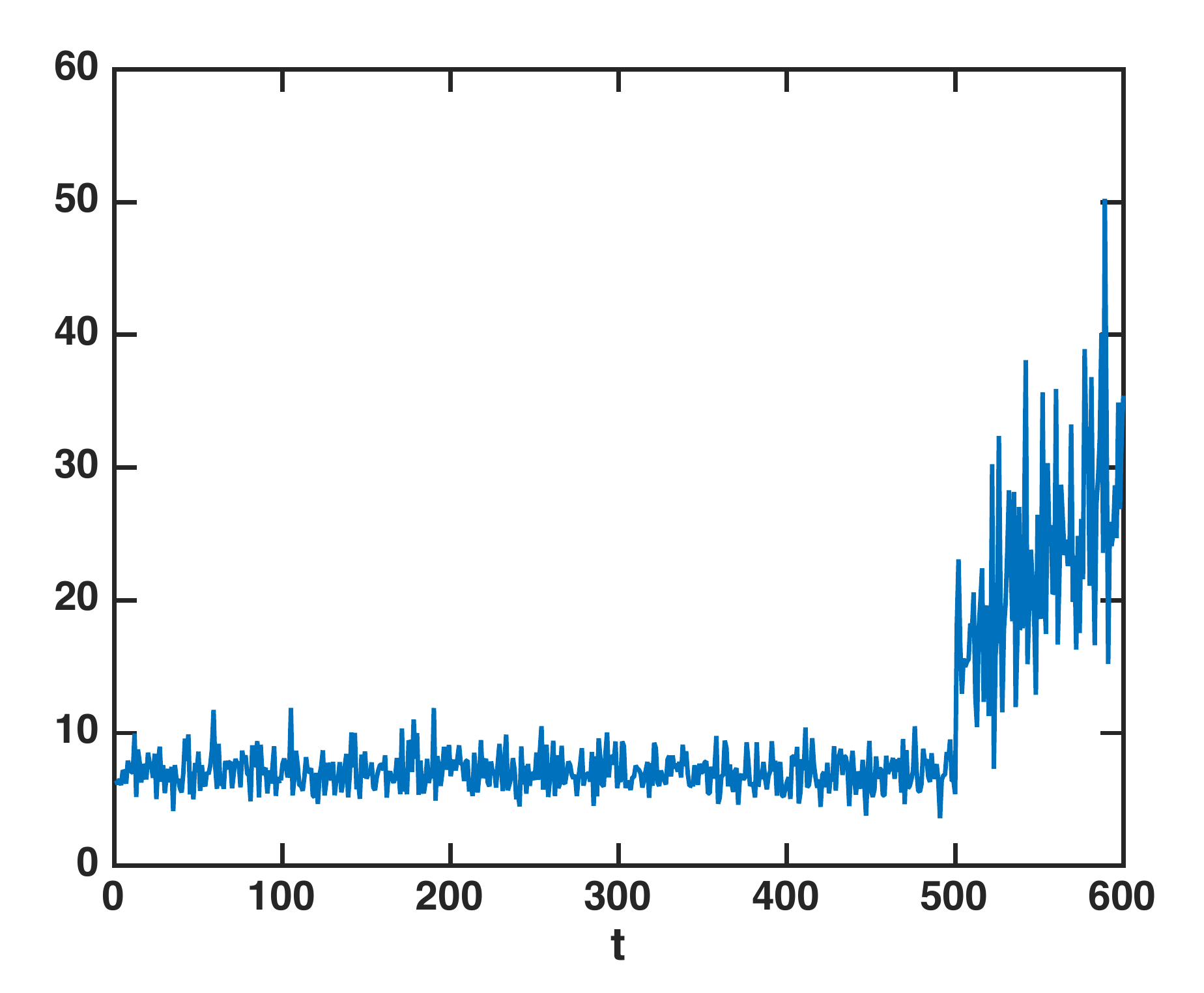}\\
(b): $\|\beta_t\|$ \\
\caption{Detection statistic $\|\beta_t\|^2$ computed using subspace tracking via Grouse \cite{GrouseGLOBALconv2015}. There is a change-point at $\kappa = 500$. In this example, $p = 100$, $s = 10$, and $\sigma_0^2 = 0.01$. This example considers missing data. Only about $70\%$ entries are observed. At each time we only randomly observe a subset of entries of $x_t$.}
\label{fig:GROUSE}
\end{figure}

There is a connection of the low-rank covariance and subspace model. Assume a rank $s$ post-change covariance matrix $\Sigma$, and its eigen-decomposion $U\Lambda U^\intercal$. Then we may expression each observation vector $x_t = U\beta_t + w_t$, where $w_t$ is a $p$-dimensional Gaussian random vector with covariance matrix $\sigma_0^2 I_p$. Before the change, $\beta_t = 0$. After the change, $\beta_t \neq 0$ and is a $s$-dimensional Gaussian random vector with covariance matrix being $\Lambda$. Hence, we may detect the low-rank change by detecting a non-zero $\beta_t$, if $U$ is known. When $U$ is unknown, we may perform an online subspace estimation from a sequence of data. 

Based on such a connection, we may develop an efficient way to compute the detection statistic online via subspace tracking.  This is related to the so-called {\it matched-subspace detection} \cite{ScharfMatchedSubspace94}, and here we further combine matched-subspace detection with online subspace estimation. Start with an initialization of the subspace $U_0$, using the sequence of observations $x_1, x_2, \ldots$, we may update the subspace using stochastic gradient descent on grassmannian manifolds, e.g., via the GROUSE algorithm \cite{GrouseGLOBALconv2015}. Then we perform projection of the next vector $U_{t-1}^\intercal x_{t}$, which gives an estimate for $\beta_{t}$. We may claim a change when either $\max_i |[\beta_t]_i|$ (mimicking the largest eigenvalue) or the norm square $\|\beta_{t}\|^2$ (mimicking the sum of the eigenvalues) becomes large. Since the subspace tracking algorithm (e.g., GROUSE) can even deal with missing data, this approach allows us to compute the detection statistic even when we can only observe a subset of entries of $x_t$ at each time. Fig. \ref{fig:GROUSE} demonstrates the detection statistic computed via GROUSE subspace tracking when only about 70$\%$ of the entries can be observed at random locations. There is a change-point at time $500$. Clearly, the detection statistic computed via subspace tracking can detect such a change by raise a large peak.

\section{Conclusion}

We have presented a sequential change-point detection procedure based on the largest eigenvalue statistic for detecting a low-rank change to the signal covariance matrix. It is related to the so-called spiked covariance model. We present a lower-bound for the average-run-length (ARL) of the procedure when there is no change, which can be used to control false alarm rate. We also present an approximation to the expected detection delay (EDD) of the proposed procedure, which characterizes the dependence of EDD on the spectral norm of the post-change covariance matrix $\Sigma$. Our theoretical results are obtained using an $\varepsilon$-net argument, which leads to a decomposition of the proposed procedure into a set of $\chi^2$-CUSUM procedures. We further present a sketching procedure, which linearly projects the original observations into  lower-dimensional sketches, and performs the low-rank change detection using these sketches. We demonstrate that when the number of sketches is sufficiently large (on the order of $p/\log s$ with $s$ being the rank of the post-change covariance), the sketching procedure can detect the change with little performance loss relative to the original procedure using full data. Finally, we present an approach to compute the detection statistic online via subspace tracking.

\section*{Acknowledgement}

The author would like to thank Dake Zhang at Tsinghua University, Lauren Hue-Seversky and Matthew Berger at Air Force Research Lab, Information Directorate for stimulating discussions, and Shuang Li at Georgia Tech for help with numerical examples. This work is partially supported by a AFRI Visiting Faculty Fellowship.

\bibliographystyle{ieeetr}
\bibliography{sample,yao_research_statement,slope_letter,yao_proposal,bib_paper,Poisson_MC,sigproc,bib,bib1,RobustChangeDetection,MOUSSE}

\end{document}